\documentclass{article}

\PassOptionsToPackage{numbers, compress}{natbib}
  \usepackage[preprint]{neurips_2026}

\usepackage[utf8]{inputenc} 
\usepackage[T1]{fontenc}    
\usepackage{hyperref}       
\hypersetup{hidelinks,pdftitle={Controlling Temporal Pseudo-Label Marginals for Stable Online Test-Time Adaptation},pdfauthor={Rani Amar and Ofir Lindenbaum}}
\usepackage{url}            
\usepackage{booktabs}       
\usepackage{amsfonts}       
\usepackage{nicefrac}       
\usepackage{microtype}      
\usepackage{xcolor}         

\usepackage{graphicx}
\usepackage{algorithm}
\usepackage[noend]{algpseudocode}
\usepackage{pgfplots}
\pgfplotsset{compat=1.18} 
\usepackage{colortbl}
\usepackage{tikz}
\usepackage[table]{xcolor}  
\usetikzlibrary{positioning, calc, arrows.meta, shapes.geometric, decorations.pathreplacing}
\definecolor{scalarBlue}{HTML}{1F77B4}
\usepackage{amsmath,amssymb} 
\usepackage[accsupp]{axessibility}  

\usepackage{orcidlink}

\definecolor{baseBlue}{RGB}{31,119,180}  
\definecolor{scalarOrange}{RGB}{255,127,14} 

\title{Controlling Temporal Pseudo-Label Marginals for Stable Online Test-Time Adaptation}

\author{%
  Rani Amar \\
  Faculty of Engineering \\
  Bar-Ilan University \\
  Ramat Gan, Israel \\
  \texttt{rani.amar@biu.ac.il} \\
  \And
  Ofir Lindenbaum \\
  Faculty of Engineering \\
  Bar-Ilan University \\
  Ramat Gan, Israel \\
  \texttt{ofir.lindenbaum@biu.ac.il} \\
}

\begin{document}

\maketitle

\begin{abstract}

Continual Test-Time Adaptation (CTTA) methods update models online using a stream of unlabeled data to overcome distribution shifts encountered at test time. However, CTTA is often unstable, as it can reinforce biases in early predictions and increasingly predict only a few classes, resulting in drift or even collapse. In this paper, we identify narrowing pseudo-label class usage as an indicator of CTTA instability and quantify this behavior. Most existing methods have attempted to address this issue by either selecting reliable adaptation samples or re-weighting the loss to encourage diverse pseudo-labels, but they do not explicitly enforce a class marginal over a temporal window. Instead, we propose Sinkhorn Class-Marginal Regularization (SCALAR), a simple plug-and-play module that controls the pseudo-label distribution over a short temporal window. SCALAR works by maintaining recent predictions in a buffer, balancing their class distribution via the Sinkhorn algorithm to match a specified target distribution, and using the balanced soft targets for adaptation. Without changing the model architecture, SCALAR maintains the per-step runtime and increases the peak allocated GPU memory by only 10 MB on ImageNet-C. Our method is particularly effective on skewed, non-stationary data with severe distribution shifts, where SCALAR significantly alleviates drift and collapse when combined with existing stabilization methods based on entropy or diversity.
Code is available at \url{https://github.com/r-isr/SCALAR}.

\end{abstract}

\section{Introduction}
\label{sec:intro}

Continual test-time adaptation (CTTA) assumes a deployment setting where a model must continuously adapt at test time under non-stationary conditions.

In CTTA, models must adapt online to a stream of unlabeled test samples that may differ in distribution from the training data and may change over time, without access to the source data~\cite{CoTTA2022}. Typically, CTTA methods adopt a teacher--student framework where an exponential moving average (EMA) of the student parameters defines the \textit{teacher} model, which generates pseudo-labels for each batch. The \textit{student} is updated online using these pseudo-labels as supervision~\cite{CoTTA2022,RMT23,RoTTA23,VIDA2024,ADMA2024}. Other methods such as TENT~\cite{Tent2021} minimize an entropy objective to directly update the BatchNorm affine parameters of a single model.

In both settings, the adaptation signal is driven by the model's own predictions. Small biases can cause the adaptation signal to focus on fewer classes even in approximately balanced data streams due to the self-training effect. This can cause the pseudo-label distribution to become less diverse and eventually collapse to a small subset of classes. The CTTA community has observed two failure modes: (1) error accumulation during test-time adaptation, ranging from \textit{drift} (gradual accumulation of errors) to \textit{collapse} (rapid accumulation of errors), and (2) \textit{forgetting} of the model's accuracy on clean data after adapting on corrupted data~\cite{CoTTA2022,RoTTA23,note2022}. Existing CTTA methods attempt to mitigate temporal instabilities by modifying local model updates. These include filtering out potentially unreliable predictions~\cite{eata2022,sar2024}, promoting pseudo-label diversity~\cite{versatile2024}, adding regularization to prevent the test-time model from deviating too far from the source model~\cite{eata2022,Universal24}, and reducing temporal correlation in adaptation batches using a memory bank~\cite{note2022,RoTTA23}. However, they do not explicitly project window-level pseudo-label class marginals toward a specified target.

We aim to control the temporal class marginals of the pseudo-labels, which form the supervision signal in CTTA. We propose Sinkhorn Class-Marginal Regularization (SCALAR), a lightweight intervention to mitigate the marginal contraction that can contribute to test-time collapse.

Concretely, SCALAR maintains a short first-in-first-out (FIFO) queue of recent teacher logits and balances their class marginals over this window using the Sinkhorn--Knopp algorithm. This generates soft labels by rescaling teacher predictions so their class marginals over the window are close to a user-specified target prior (uniform by default). This procedure approximates a constrained KL projection of teacher predictions to satisfy a desired temporal class marginal~\cite{sinkhorn1964,cuturi2013}.

SCALAR is not intended to recover the test-time class prior but to stabilize the supervision signal observed by the student. While providing the correct class prior can reduce corruption errors, SCALAR with a uniform prior still provides substantial stabilization even in skewed settings. However, SCALAR with a uniform prior can slightly hinder the model's ability to specialize to target classes if the target distribution consists of only a subset of source classes and the model is already stably adapting to them.

We find SCALAR's stabilization benefit is highly state-dependent: when class usage is stable, SCALAR often has a smaller effect. But as class usage contracts and errors increase, SCALAR can counteract this contraction. We evaluate this behavior across balanced and skewed streams, fixed and dynamic skew schedules, different adaptation speeds, both teacher--student and entropy minimization settings, and backbone architectures like ViT-B and ResNet-50.

Since SCALAR is a window-level intervention, it complements other methods that stabilize updates on a per-sample (e.g., filtering, loss weighting) or per-parameter level (regularization). We demonstrate SCALAR's complementarity to the EATA method~\cite{eata2022}, which uses an entropy- and diversity-aware, history-dependent approach to select samples for updating. Despite adaptive sample selection, EATA does not explicitly constrain class marginals of the supervision signal. Applying SCALAR on top of EATA further reduces drift and mitigates collapse.

\paragraph{Contributions.}
Our contributions are threefold:

\begin{itemize}
\item \textbf{Temporal control of pseudo-labels.}
We propose SCALAR, which uses Sinkhorn balancing over a short queue of recent predictions to bring their class marginals close to a specified target prior.

\item \textbf{When SCALAR helps.}
We track the effective number of pseudo-label classes to study when marginal balancing reduces drift and collapse. We also show that a uniform prior can hinder specialization when only a subset of source classes is present, and adaptation is already stable.

\item \textbf{Stabilization across adaptation methods.}
We show that SCALAR reduces drift and collapse under skewed streams and aggressive adaptation in both teacher--student and entropy-minimization settings, including EATA. Our ImageNet-C measurements show no measurable increase in runtime and a 10\,MB increase in peak allocated GPU memory.

\end{itemize}

\section{Related Work}
\label{sec:related}

\paragraph{Test-time and continual test-time adaptation.}

Test-time adaptation (TTA) updates a model at inference using unlabeled test samples~\cite{TTT20,AdaBN16}. TENT~\cite{Tent2021} does this by minimizing prediction entropy with respect to the BatchNorm affine parameters. Continual TTA (CTTA) considers streams whose distribution changes over time~\cite{CoTTA2022}. Many CTTA methods use a teacher--student framework, where an EMA teacher generates pseudo-labels, and the student learns from them online~\cite{CoTTA2022,RMT23,RoTTA23,VIDA2024,ADMA2024}. Related source-free domain adaptation work constructs prototype-based pseudo-labels through probabilistic reasoning in concept space for concept bottleneck models~\cite{eisenberg2026probabilistic}. Self-supervised learning methods, including contrastive learning, teacher--student learning, and masked image modeling~\cite{SimCLR20,MoCo20,BYOL20,DINO2021,MAE2022,IJEPA2023}, have also informed auxiliary objectives for adaptation without labels~\cite{ContrastiveTTA22,ADMA2024}.

In multi-view clustering, COPER~\cite{eisenberg2025coper} learns cluster assignments from pseudo-labels that are consistent across views.

SeLa~\cite{SeLa2020} and SwAV~\cite{SWAV2020} use Sinkhorn balancing to avoid degenerate cluster assignments. For unsupervised permutation learning, the work in \cite{eisenberg2026permutation} introduced entropy-adaptive Gumbel--Sinkhorn normalization to improve assignment stability. ReMixMatch~\cite{ReMixMatch2020} aligns pseudo-labels with the labeled-data class distribution using a moving average of predictions over recent batches. Rozner et al.~\cite{rozner2024domain} use prediction-based label smoothing to discourage collapse into frequently predicted clusters when learning from unlabeled source domains for generalization to unseen domains. SCALAR instead applies a joint Sinkhorn projection over a short prediction window during CTTA, using a specified target prior without source data.

\noindent \textbf{Class-prior shift and temporally correlated streams.}
Temporally correlated inputs and class-ordered streams can increase error accumulation and cause prediction collapse in CTTA~\cite{RoTTA23,RMT23,Universal24,note2022}. Universal TTA~\cite{Universal24} examines these failures under class imbalance and class-ordered streams, while NOTE~\cite{note2022} studies adaptation to temporally correlated samples.

Existing methods address these failures by changing either the predictions or the samples used for adaptation. Universal TTA~\cite{Universal24} reweights predictions using a class-prior estimate from the current batch. LAME~\cite{LAME2022} adjusts predictions using similarities between samples in feature space, without updating the model weights. EATA~\cite{eata2022} updates the model only on samples selected for reliability and diversity. DELTA~\cite{DELTA2023} addresses class bias through online loss reweighting, while TRIBE~\cite{TRIBE2024} uses balanced normalization and an anchored self-training loss.

SCALAR acts on the pseudo-labels used to update the model. It collects recent predictions in a short queue and balances their class marginals toward a specified prior. The balanced targets then provide supervision for adaptation, limiting the concentration of pseudo-label mass on a few classes.

\begin{figure}[t]
\centering
\includegraphics[width=\linewidth]{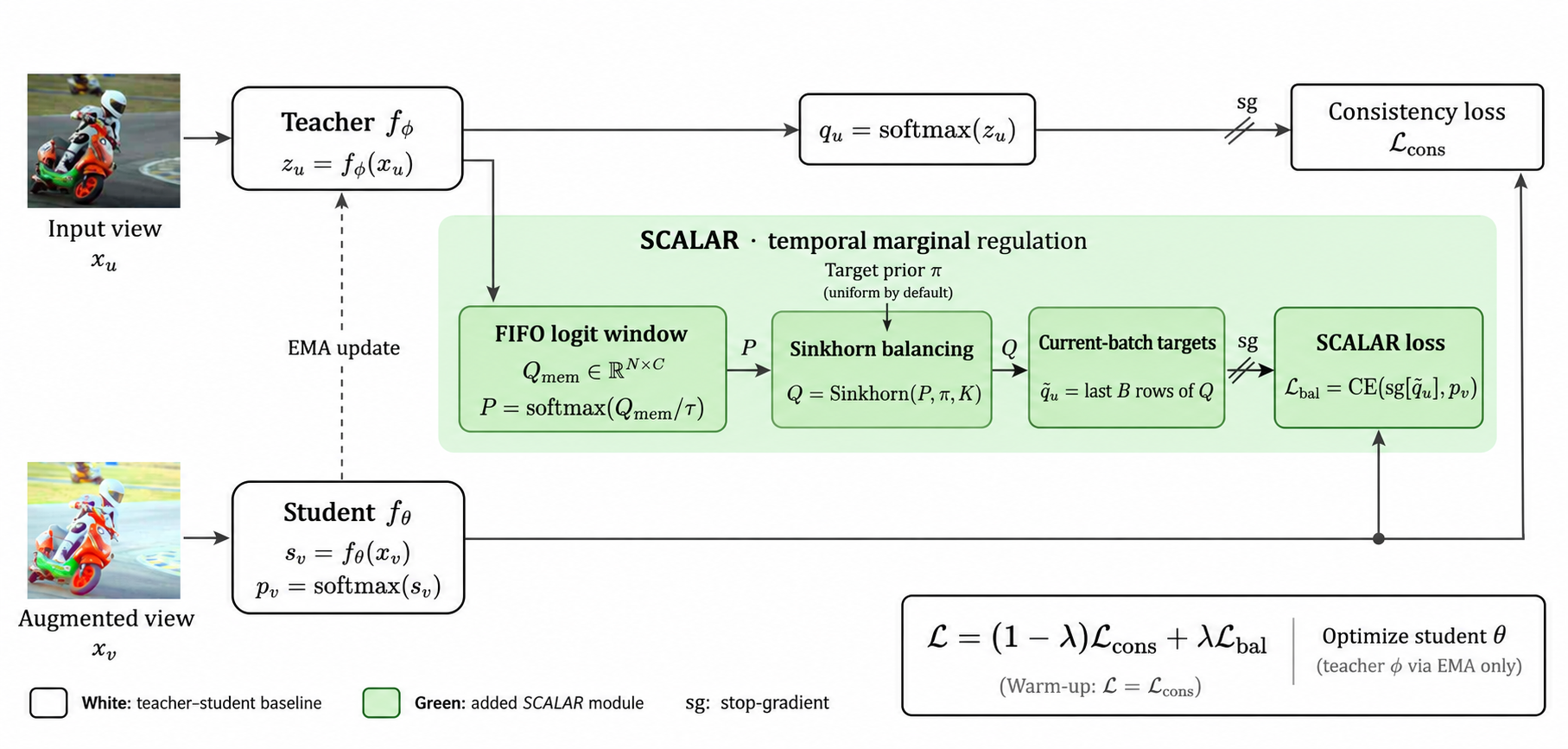}
\caption{
\textbf{Overview of the SCALAR framework.}
White blocks denote the baseline teacher--student CTTA pipeline; green blocks denote the added SCALAR module.
Given an input view $x_u$ for the teacher branch and an augmented view $x_v$ for the student branch,
the teacher produces logits $z_u$ and probabilities $q_u=\mathrm{softmax}(z_u)$.
Recent teacher logits are stored in a FIFO window and converted to probabilities $P$, which are balanced toward a target prior $\pi$ via Sinkhorn to produce balanced targets $\tilde{q}_u$.
The student produces logits $s_v$ and probabilities $p_v$, and is optimized using a mixture of the baseline consistency loss and the SCALAR loss based on $\tilde{q}_u$.
Gradients are stopped through teacher-derived targets, marked by $\mathrm{sg}$, and teacher parameters are updated via EMA.
Here $\mathcal{L}_{\mathrm{cons}}\equiv\mathcal{L}_{\mathrm{consistency}}$ and $\mathcal{L}_{\mathrm{bal}}=\mathrm{CE}(\mathrm{sg}[\tilde{q}_u],p_v)$.
}
\label{fig:scalar_overview}
\end{figure}

\section{Method}
\label{sec:method}

We first describe the teacher--student baseline, then introduce SCALAR as a step that balances the pseudo-labels used to update the student. SCALAR collects recent predictions in a short queue so that class usage is controlled across several batches. The architecture, optimizer, and update schedule remain the same. Fig.~\ref{fig:scalar_overview} shows the full procedure.

Let $f_{\theta_0}:\mathcal{X}\rightarrow\mathbb{R}^C$ be a $C$-class classifier. In the CTTA setting, the model adapts online on an unlabeled stream $\{x_t\}_{t\ge 1}$ drawn from a potentially shifting distribution, without access to source data. We denote logits by $f_\theta(x)\in\mathbb{R}^C$ and probabilities by $\mathrm{softmax}(f_\theta(x))$. At each step, the model makes predictions and may use them to perform one adaptation update.

\subsection{Teacher--Student CTTA Baseline}
\label{sec:baseline}
Our main experiments compare the same teacher--student baseline with and without SCALAR. The student $f_\theta$ learns from the teacher's predictions through gradient descent. The teacher $f_\phi$ follows the student through an exponential moving average (EMA) of its parameters. We describe these two updates below; augmentations and hyperparameters are given in App.~\ref{app:implementation}.

\noindent \textbf{Two-view consistency.}
For each unlabeled input $x_u$, we form an augmented view $x_v=\mathcal{T}(x_u)$, where $\mathcal{T}$ denotes a stochastic data augmentation operator. The teacher predicts from the input without further augmentation, while the student predicts from the augmented view:
\begin{align*}
z_u &= f_\phi(x_u), \qquad q_u = \mathrm{softmax}(z_u), \\
s_v &= f_\theta(x_v), \qquad p_v = \mathrm{softmax}(s_v).
\end{align*}
The student minimizes a symmetric consistency loss:
\begin{equation*}
\mathcal{L}_{\mathrm{consistency}}
= \tfrac{1}{2}\!\left[
\mathrm{CE}(\mathrm{sg}[q_u], p_v)
+ \mathrm{CE}(p_v, \mathrm{sg}[q_u])
\right],
\end{equation*}
where $\mathrm{sg}[\cdot]$ denotes stop-gradient, and
$\mathrm{CE}(a,b)=-\sum_{c=1}^{C} a_c \log b_c$
denotes the cross-entropy between distributions $a$ and $b$. This loss encourages the student to agree with the teacher despite the augmentation. The teacher predictions are held fixed during the student update.

\paragraph{EMA teacher.}
After each student update, the teacher's parameters are updated as an exponential moving average with momentum coefficient $\alpha\in[0,1)$:
\begin{equation*}
\phi \leftarrow \alpha\,\phi + (1-\alpha)\,\theta .
\end{equation*}
A larger $\alpha$ makes the teacher follow the student more slowly, reducing the effect of any single update. However, averaging the parameters does not directly control which classes receive pseudo-label mass. SCALAR adds this control over a window of recent predictions. In our ViT baseline, attention and the remaining parameters use separate EMA rates (App.~\ref{app:dual_vs_single_ema}); $\alpha$ denotes the rate for the corresponding parameters.

\subsection{SCALAR: Class-Marginal Balancing via a Logit Queue}
\label{sec:scalar-framework}

SCALAR balances predictions over a window of $N$ samples, including the current batch of size $B$. We use a temperature $\tau>0$ to convert logits to probabilities and $K$ Sinkhorn iterations to balance them toward a target prior $\pi\in\Delta^{C-1}$, which is uniform by default. The weight $\lambda\in[0,1]$ controls the contribution of the balanced targets to the student loss.

\noindent \textbf{Logit queue.}
We maintain a FIFO queue $Q_\mathrm{mem}\in\mathbb{R}^{N\times C}$
containing the most recent teacher logits. At each step, the oldest $B$ logits are removed, and the current logits $z_u$ are appended. The queue is converted to a probability matrix via a row-wise softmax:
\begin{equation*}
P=\mathrm{softmax}(Q_\mathrm{mem}/\tau)\in\mathbb{R}^{N\times C},
\qquad
P_{i,j}=
\frac{\exp((Q_\mathrm{mem})_{i,j}/\tau)}
{\sum_{c=1}^{C}\exp((Q_\mathrm{mem})_{i,c}/\tau)}.
\end{equation*}
The column sums $P^\top\mathbf{1}_N$ give the total probability mass assigned to each class in the window, where $\mathbf{1}_N$ is a vector of ones. Dividing by $N$ gives the average class distribution. Using several recent batches allows us to balance this distribution without requiring each small batch to represent all classes.
\begin{algorithm}[t]
\small
\caption{\textbf{SCALAR Online Update.}}
\label{alg:scalar}
\begin{algorithmic}[1]
\State \textbf{Input:} batch $x_u$, teacher $f_\phi$, student $f_\theta$,
queue $Q_\mathrm{mem}$ (capacity $N$), temperature $\tau$, iterations $K$,
prior $\pi$, weight $\lambda$
\State $x_v\gets\mathcal{T}(x_u)$;\quad
$z_u\gets f_\phi(x_u)$;\quad
$q_u\gets\mathrm{softmax}(z_u)$;\quad
$p_v\gets\mathrm{softmax}(f_\theta(x_v))$
\If{$|Q_\mathrm{mem}|=N$}
    \State Update queue (dequeue oldest $B$, enqueue $z_u$)
    \State $P\gets\mathrm{softmax}(Q_\mathrm{mem}/\tau)$
    \State $Q\gets\mathrm{SinkhornBalance}(P,\pi,K)$
    \State $\tilde{q}_u\gets$ last $B$ rows of $Q$
    \State $\mathcal{L}\gets(1-\lambda)\mathcal{L}_{\mathrm{consistency}}
    +\lambda\,\mathrm{CE}(\mathrm{sg}[\tilde{q}_u],p_v)$
\Else
    \State $Q_\mathrm{mem}\gets\mathrm{Enqueue}(Q_\mathrm{mem},z_u)$
    \State $\mathcal{L}\gets\mathcal{L}_{\mathrm{consistency}}$ \Comment{warm-up}
\EndIf
\State $\theta\gets\theta-\eta\nabla_\theta\mathcal{L}$;\quad
$\phi \leftarrow \alpha\,\phi + (1-\alpha)\,\theta$
\end{algorithmic}
\end{algorithm}

\paragraph{Sinkhorn--Knopp balancing.}
Given the target prior $\pi$, we seek a matrix $Q$ whose rows sum to one and whose column sums match $N\pi$.
Starting from $Q^{(0)}=P$, we iterate:
\begin{equation}
\tilde{Q}^{(k)}=
\mathrm{Diag}(Q^{(k)}\mathbf{1}_C)^{-1}Q^{(k)},
\qquad
Q^{(k+1)}=
\tilde{Q}^{(k)}
\,\mathrm{Diag}\!\left(
\frac{N\pi}{\tilde{Q}^{(k)\top}\mathbf{1}_N}
\right),
\label{eq:sinkhorn}
\end{equation}
followed by a final row normalization. Here $\mathrm{Diag}(\cdot)$ forms a diagonal matrix from a vector. Row normalization makes each row a probability distribution, while column scaling adjusts the total mass assigned to each class. After a finite number of iterations, the column sums approximately match $N\pi$. This procedure approximates the constrained KL projection of $P$~\cite{sinkhorn1964,cuturi2013}. Since classes are rescaled by different factors, their ordering within a sample can change. We use the last $B$ rows of $Q$ as the balanced targets $\tilde{q}_u$ for the current batch; past predictions provide context for balancing.

\noindent \textbf{Combined objective.}
We update the student using a mixture of the consistency loss and the loss on balanced targets $\tilde{q}_u$:
\begin{equation}
\mathcal{L}
=(1-\lambda)\mathcal{L}_{\mathrm{consistency}}
+\lambda\,\mathrm{CE}(\mathrm{sg}[\tilde{q}_u],\,p_v),
\label{eq:total-loss}
\end{equation}
where $\lambda$ determines how strongly the balanced targets influence the student relative to the original consistency loss. As above, we stop gradients through the targets and update the teacher only via EMA. While the queue fills, we train with $\mathcal{L}_{\mathrm{consistency}}$ alone.

Algorithm~\ref{alg:scalar} summarizes the online update procedure.

\noindent \textbf{Computational cost and wall-clock overhead.}
SCALAR adds a Sinkhorn balancing step over the logit queue. Each update costs $\mathcal{O}(KNC)$ time and $\mathcal{O}(NC)$ memory, where $N$ is the queue length, $C$ the number of classes, and $K$ the number of Sinkhorn iterations. For ImageNet-C ($C{=}1000$, $N{=}1024$, $K{=}5$), this is about $0.02$ GFLOPs and a $\sim$4\,MB fp32 queue. Measured per-step time is unchanged within noise (0.1063s $\rightarrow$ 0.1062s), and peak GPU memory increases by 10\,MB allocated / 20\,MB reserved. Full overhead details are provided in App.~\ref{app:overhead}.

\begin{figure}[t]
    \centering
    \includegraphics[width=\linewidth]{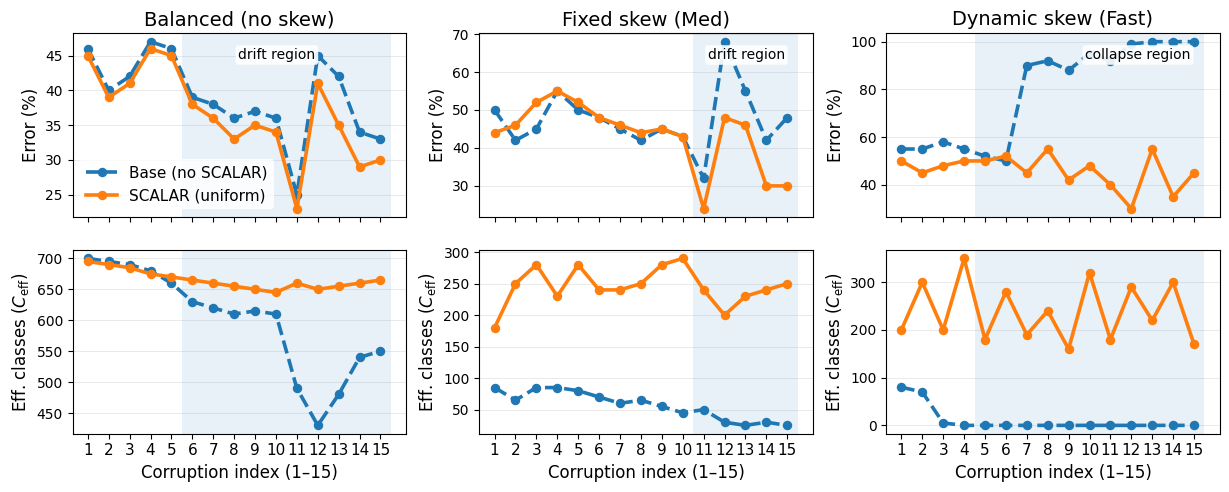}
    \caption{
    \textbf{Temporal contraction across adaptation regimes.}
    Temporal evolution of error (top) and effective pseudo-label classes $C_\mathrm{eff}$ (bottom)
    in a teacher--student continual test-time adaptation setup with a ViT-B/16 backbone
    (ImageNet-C, severity 5, single-pass protocol).
    Here, $C_\mathrm{eff}$ denotes the effective number of classes in the teacher
    pseudo-label distribution.
    \textbf{Left:} Balanced stream (no skew) exhibits gradual contraction and late-stage drift.
    \textbf{Middle:} Fixed skew with rotating dominant classes (medium adaptation) results
    in stronger drift.
    \textbf{Right:} Dynamic skew with fast adaptation induces rapid contraction leading to collapse.
    Across regimes, SCALAR stabilizes class usage and mitigates the associated error growth.
    }
    \label{fig:dynamics_across_regimes}
\end{figure}

\begin{table*}[t]
\centering
\caption{
\textbf{Balanced corruption benchmarks (severity 5).}
Average top-1 error (\%) over all 15 corruptions on CIFAR-10-C, CIFAR-100-C, and ImageNet-C with ViT-B/16 backbones. \emph{SCALAR} denotes our teacher--student CTTA pipeline with the FIFO logit queue and Sinkhorn balancing; \emph{Baseline w/o SCALAR} is the same pipeline without this module.
SCALAR uses a uniform class prior. For SCALAR and the teacher--student baseline, results are averaged over 5 seeds (std. $\approx 0.2$); prior-method numbers use their published settings, which can differ in initialization and batch size.
}
\label{tab:ctta_iid_all}
\setlength{\tabcolsep}{4pt}
\renewcommand{\arraystretch}{1.05}
\begin{tabular}{lccc}
\toprule
Method & CIFAR-10-C & CIFAR-100-C & ImageNet-C \\
\midrule
Source-only                         & 28.1 & 35.4 & 55.8 \\
TENT~\cite{Tent2021}                & 23.5 & 32.1 & 51.0 \\
CoTTA~\cite{CoTTA2022}              & 24.6 & 34.8 & 54.8 \\
VDP~\cite{VDP2023}                  & 24.1 & 32.0 & 50.0 \\
VIDA~\cite{VIDA2024}                & 20.7 & 27.3 & 43.4 \\
Versatile~\cite{versatile2024}      & 8.9  & 24.0 & 42.7 \\
Universal~\cite{Universal24}        & --   & --   & 45.0 \\
ADMA~\cite{ADMA2024}                & 12.6 & 26.4 & 42.5 \\
REM~\cite{REM2025}                  & 9.4  & 23.4 & 39.2 \\
\midrule
\rowcolor{orange!15}
\textbf{SCALAR (ours)}              & \textbf{8.1} & \textbf{23.2} & \textbf{36.1} \\
\midrule
Baseline (w/o SCALAR)   
                                   & 10.3 & 24.7 & 38.1 \\
\bottomrule
\end{tabular}
\end{table*}

\begin{table}[t]
\centering
\caption{
\textbf{Fixed-skew continual streams across adaptation speeds (ImageNet-C, severity 5).} Each cell reports average corruption error and post-stream clean error (\%), mean\,$\pm$\,SD over three runs.
Adaptation speed is controlled by the learning rate and EMA momentum (App.~\ref{app:exp_settings}).
\emph{SCALAR-U} uses a uniform class prior; \emph{SCALAR-C} uses the class prior of each corruption segment (oracle reference).
At high speed, the baseline collapsed in 2/3 runs; all SCALAR runs remained stable.
}
\label{tab:imagenet_skewed_speeds_compact_clean}
\providecommand{\sd}[1]{\,{\scriptsize$\pm$#1}}
\setlength{\tabcolsep}{4pt}
\renewcommand{\arraystretch}{1.12}
\small
\begin{tabular}{lcc|cc|cc}
\toprule
& \multicolumn{2}{c|}{High speed}
& \multicolumn{2}{c|}{Medium speed}
& \multicolumn{2}{c}{Low speed} \\
Method & Avg Err & Clean Err & Avg Err & Clean Err & Avg Err & Clean Err \\
\midrule
Baseline (w/o SCALAR)
& 86.8\sd{14.4} & 76.4\sd{40.6}
& 48.3\sd{4.4} & 25.2\sd{5.9}
& \textbf{47.5}\sd{0.6} & 17.4\sd{0.3} \\
\rowcolor{orange!15}
\textbf{SCALAR-U}
& 48.1\sd{1.1} &  \textbf{21.1}\sd{0.5}
& 42.9\sd{0.7} & 18.2\sd{0.4}
& 49.1\sd{0.4} & 17.3\sd{0.1} \\
\rowcolor{orange!15}
\textbf{SCALAR-C}
& \textbf{46.1}\sd{1.0} & 22.2\sd{0.4}
& \textbf{42.5}\sd{0.4} & \textbf{17.7}\sd{0.3}
& 48.2\sd{0.8} & \textbf{17.1}\sd{0.1} \\
\bottomrule
\end{tabular}
\end{table}

\section{Experiments}
\label{sec:experiments}


Our experiments address three questions:

\begin{itemize}

\item \textbf{(Q1) State-dependent stabilization.}
When does SCALAR reduce drift and collapse? How does its benefit change as pseudo-label class usage contracts? Does it also improve predictive calibration?

\item \textbf{(Q2) Choice of target prior.}
When the test-time class prior is unknown, does a uniform target stabilize adaptation? How much does an accurate prior help, and when does a uniform target hinder specialization?

\item \textbf{(Q3) Use with other adaptation methods.}
Does SCALAR improve methods that already select samples for reliability and diversity? Does it also help TENT with a ResNet backbone?

\end{itemize}

\subsection{Setup}
\label{sec:setup}

\noindent \textbf{Protocol.}
Unless otherwise stated, experiments follow the standard continual test-time adaptation (CTTA) protocol~\cite{CoTTA2022}: a pretrained model adapts online to a single-pass stream of unlabeled test data, with no access to source data or ground-truth labels during adaptation. We apply this protocol to both standard corruption streams and controlled label-skewed streams constructed from ImageNet-C.

\noindent \textbf{Datasets and streams.}
We use ImageNet-C and CIFAR-10/100-C~\cite{Corrupt2019} at severity~5. These benchmarks contain corrupted versions of the original validation images, grouped into 15 corruption types such as noise, blur, weather, and digital artifacts, which are processed sequentially to form a continual test stream. For the standard balanced benchmarks, we use the original corruption streams and report average top-1 error across all corruptions.

To study sustained prior shift, we construct fixed- and dynamic-skew ImageNet-C streams by resampling the class composition within each corruption segment; full details are provided in App.~\ref{app:skew}. In both cases, classes are divided into head and tail groups, and skew is induced by oversampling head-class examples while subsampling tail-class examples. In the fixed-skew setting, each segment contains 50 head classes and 950 tail classes; each head class is sampled to 100 examples, while each tail class contributes one example. Thus, the per-class head--tail ratio is 100:1, with roughly 84\% of the segment mass assigned to head classes. The skew strength is fixed across the stream, while the dominant class identities are deterministically reshuffled across corruption segments. In the dynamic-skew setting, the head-set size and per-class oversampling strength both vary across segments, producing head--tail ratios ranging from 25:1 to 250:1 while keeping the aggregate head mass roughly comparable. All methods  are evaluated on streams with the same skew configuration. Long-horizon persistent-skew and true-subset variants are defined with the corresponding experiments; full construction details are provided in App.~\ref{app:skew}.

\noindent \textbf{Models.}
Our primary experiments use a ViT-B/16 backbone within a teacher--student framework.
We also add SCALAR to EATA~\cite{eata2022} with ViT-B/16 and to TENT~\cite{Tent2021} with a ResNet-50 backbone.

\noindent \textbf{SCALAR settings.}
Unless otherwise stated, we use $N=1024$, $\tau=0.1$, and $\lambda=0.5$; for EATA, whose entropy filter retains only sharply peaked predictions, we use $\tau=0.7$ and $N=4096$. Learning rates and EMA momenta are listed in App.~\ref{app:exp_settings}.

\noindent \textbf{Metrics.}
We report:
(i) top-1 corruption error (lower is better),
(ii) post-stream clean error, measured on clean data after adaptation to quantify forgetting,
and (iii) the effective number of pseudo-label classes, $C_\mathrm{eff}(t)$, which we use only as a label-free diagnostic of temporal class-marginal contraction, not as an accuracy measure:
\begin{equation*}
C_{\mathrm{eff}}(t) = \exp\!\Big( - \sum_{y=1}^{C} \bar{p}_t(y)\,\log \bar{p}_t(y) \Big),
\end{equation*}
where $\bar{p}_t$ is the window-averaged pseudo-label distribution (before Sinkhorn balancing). $C_{\mathrm{eff}}(t)=C$ under uniform class usage and decreases as probability mass concentrates on fewer classes. Concentration can be appropriate when the stream contains only a few classes, and diverse predictions can still be wrong. We therefore examine $C_{\mathrm{eff}}(t)$ alongside error when studying drift and collapse.

\noindent \textbf{Baselines and variants.}
\textbf{Baseline} denotes our teacher--student CTTA pipeline without SCALAR's Sinkhorn marginal-balancing module.
\textbf{SCALAR} or \textbf{SCALAR full framework} denotes the same teacher--student pipeline with the SCALAR module.
When SCALAR is added to an existing method, we write \textbf{Method+SCALAR}, e.g., EATA+SCALAR or TENT+SCALAR, meaning that only the SCALAR marginal-regulation module is added to that method.
Unless otherwise stated, SCALAR uses a uniform class prior. We write \textbf{SCALAR-U} for this variant and \textbf{SCALAR-C} for the oracle-prior variant; the target is the segment prior or aggregate stream prior, as specified for each experiment.

\subsection{State-Dependent Stabilization}
\label{sec:q1}

We first examine how pseudo-label class usage changes during adaptation. We track $C_{\mathrm{eff}}(t)$, defined in Sec.~\ref{sec:setup}, together with error to see when class contraction accompanies a loss of accuracy.

Fig.~\ref{fig:dynamics_across_regimes} shows the evolution of corruption error and $C_{\mathrm{eff}}(t)$ across balanced, fixed-skew, and dynamic-skew regimes. Error tends to increase as $C_{\mathrm{eff}}(t)$ decreases. Balanced streams show mild gradual contraction, fixed skew strengthens this effect, and dynamic skew with rapid adaptation leads to sharp contraction and collapse.

On standard balanced corruption benchmarks (Tab.~\ref{tab:ctta_iid_all}), where contraction is mild, SCALAR gives moderate improvements (e.g., $38.1\%\!\rightarrow\!36.1\%$ on ImageNet-C). Under stronger drift or collapse, SCALAR reduces the decline in $C_{\mathrm{eff}}(t)$ and the associated rise in error. Its benefit is larger in these unstable regimes.

\subsection{Uniform Targets under Class Skew}
\label{sec:q2}

A uniform target may differ substantially from the class distribution in a skewed stream. We test whether it can still stabilize adaptation, then compare it with the correct prior to measure the cost of this mismatch.

\noindent \textbf{Uniform prior in skewed streams.}
Tab.~\ref{tab:imagenet_skewed_speeds_compact_clean} reports results on fixed-skew ImageNet-C streams across adaptation speeds, controlled by the learning rate and EMA momentum. In high-speed regimes, where the baseline collapses in 2/3 runs, SCALAR-U reduces the corruption error from $86.8\%$ to $48.1\%$ and the post-stream clean error from $76.4\%$ to $21.1\%$. It also improves corruption and clean performance at medium speed. At low speed, however, corruption error increases from $47.5\%$ to $49.1\%$, while clean error changes from $17.4\%$ to $17.3\%$, consistent with over-regularization when adaptation is already stable. Thus, the benefit depends on whether adaptation is becoming unstable, even though the class skew is the same at all three speeds. A target-sharpening baseline also collapses in 2/3 runs
at high speed (App.~\ref{app:sharpening}).

\noindent \textbf{Benefit of an accurate prior.}
To quantify the benefit of accurate prior knowledge, we compare against \textbf{SCALAR-C}, which uses the empirical prior of each corruption segment as an oracle reference. Across the fixed-skew regimes, the correct prior further reduces corruption error relative to the uniform prior. At high speed, both targets prevent collapse in these runs. Knowing the prior improves accuracy further, but is not required for the observed stabilization.

\begin{table}[t]
\centering
\begin{minipage}[t]{0.42\linewidth}
\centering
\caption{\textbf{Calibration under aggressive adaptation}
(ViT-B/16, fixed-skew ImageNet-C, high speed as in Tab.~\ref{tab:imagenet_skewed_speeds_compact_clean}, single pass, averaged over three runs). The baseline collapsed in 2/3 runs.}
\label{tab:q3_controls}
\footnotesize
\setlength{\tabcolsep}{4pt}
\begin{tabular}{lcc}
\toprule
Method & ECE $\downarrow$ & NLL $\downarrow$ \\
\midrule
Baseline & 0.73 & 13.37 \\
\rowcolor{orange!15}
SCALAR & \textbf{0.07} & \textbf{2.15} \\
\bottomrule
\end{tabular}
\end{minipage}\hfill
\begin{minipage}[t]{0.55\linewidth}
\centering
\caption{\textbf{EATA under dynamic label skew}
(ImageNet-C, severity~5, ViT-B/16). Corruption / post-stream clean error (\%),
mean\,$\pm$\,SD over three runs (stable) and eight runs (aggressive). Under aggressive adaptation, EATA collapses in 4/8 runs, whereas all SCALAR runs remain stable.}
\label{tab:short_eata_scalar_lr}
\footnotesize
\setlength{\tabcolsep}{3pt}
\begin{tabular}{lccc}
\toprule
Method & Avg Err & Clean Err & Coll. \\
\midrule
\multicolumn{4}{l}{\emph{Stable}} \\
EATA & 50.8{\tiny$\pm$2.4} & 18.8{\tiny$\pm$0.6} & 0/3 \\
\rowcolor{orange!15}
\,+SCALAR-U &  \textbf{49.6}{\tiny$\pm$2.1} & \textbf{18.0}{\tiny$\pm$0.4} & 0/3 \\
\midrule
\multicolumn{4}{l}{\emph{Aggressive}} \\
EATA & 77.0{\tiny$\pm$11.3} & 67.0{\tiny$\pm$37.0} & 4/8 \\
\rowcolor{orange!15}
\,+SCALAR-U & \textbf{57.0}{\tiny$\pm$4.6} & \textbf{21.1}{\tiny$\pm$1.0} & \textbf{0/8} \\
\bottomrule
\end{tabular}
\end{minipage}
\end{table}

\subsection{Calibration}
\label{sec:mechanism_calibration}

We also examine whether SCALAR improves the probabilities assigned to predictions. In a separate single-pass evaluation under aggressive adaptation, SCALAR reduces expected calibration error (ECE) from $0.73$ to $0.07$ and negative log-likelihood (NLL) from $13.37$ to $2.15$ (Tab.~\ref{tab:q3_controls}). These results show improved calibration in this setting, in addition to the reductions in top-1 error reported above.

\subsection{Adding SCALAR to EATA}
\label{sec:eata_complementarity}

We next test whether SCALAR helps when the adaptation method already selects samples for reliability and diversity. EATA uses prediction entropy and a running estimate of past predictions to select samples for updating. We add SCALAR to control the class marginals of the supervision signal and evaluate both methods under dynamic skew (Tab.~\ref{tab:short_eata_scalar_lr}). In all EATA experiments, with or without SCALAR, the Fisher regularizer is disabled.

At EATA's stable operating point, adding SCALAR reduces average error from $50.8\%$ to $49.6\%$. However, EATA is sensitive to larger learning rates: under aggressive adaptation its average error rises to $77.0\%$ and it collapses in four of eight runs. SCALAR prevents collapse in all runs, reducing average error to $57.0\%$ and keeping clean error at $21.1\%$, close to its stable-regime level. Sample selection alone does not prevent collapse at this learning rate; balancing the targets provides additional stabilization.

\noindent \textbf{Long-horizon persistent skew.}
The dynamic-skew stream changes which classes dominate. To test whether the benefit persists when the imbalance lasts longer, we keep the same 50 dominant classes throughout five passes (Tab.~\ref{tab:long_horizon_persistent_skew}). EATA progressively degrades, from $58.1\%$ error in the first pass to $78.6\%$ in the fifth, and collapses in one of three runs. EATA+SCALAR-U remains stable in all runs, reaching $48.7\%$ in the fifth pass and reducing final clean error from $52.3\%$ to $21.5\%$. Using the aggregate stream prior further reduces overall error from $50.0\%$ to $43.9\%$, whereas the stability benefit is already obtained with a uniform target.

\noindent \textbf{Boundary case: true class absence.}
In the persistent-skew experiment, all classes still appear. We now remove the tail classes entirely: only the same 50 classes appear over five passes, and the remaining 950 are absent (Tab.~\ref{tab:true_subset}). A uniform target assigns mass to these absent classes, so it may hinder specialization to the observed subset. Under conservative adaptation, EATA specializes to the observed classes, whereas EATA+SCALAR-U improves less, indicating mild over-regularization; the aggregate stream prior (SCALAR-C) recovers this specialization. Under aggressive adaptation, EATA collapses in one of two runs, whereas EATA+SCALAR-U remains stable but does not improve across passes. SCALAR-C again improves steadily. The uniform target prevents collapse in this setting, but limits specialization to the classes that are present.

\begin{table}[t]
\centering
\begin{minipage}[t]{0.52\linewidth}
\centering
\caption{\textbf{Long-horizon persistent skew} (\mbox{ImageNet-C}, five passes,
$>$400K images, three runs). The same 50 classes dominate throughout
(100 samples each, one per remaining class, per corruption).
Error (\%), mean\,$\pm$\,SD.
SCALAR-C uses the aggregate class prior of the stream (gray, oracle reference). Bold: best non-oracle result.}
\label{tab:long_horizon_persistent_skew}
\footnotesize
\setlength{\tabcolsep}{2.5pt}
\begin{tabular}{lccccc}
\toprule
Method & Pass 1 & Pass 5 & Overall & Final clean & Coll. \\
\midrule
EATA & 58.1 & 78.6 & 66.5{\tiny$\pm$12.9} & 52.3{\tiny$\pm$41.3} & 1/3 \\
\rowcolor{orange!15}
\,+SCALAR-U & \textbf{49.3} & \textbf{48.7} & \textbf{50.0}{\tiny$\pm$2.7} & \textbf{21.5}{\tiny$\pm$1.4} & \textbf{0/3} \\
\textcolor{gray}{\,+SCALAR-C} & \textcolor{gray}{47.9} & \textcolor{gray}{43.3} & \textcolor{gray}{43.9{\tiny$\pm$1.7}} & \textcolor{gray}{21.4{\tiny$\pm$0.8}} & \textcolor{gray}{0/3} \\
\bottomrule
\end{tabular}
\end{minipage}\hfill
\begin{minipage}[t]{0.45\linewidth}
\centering
\caption{\textbf{Long-horizon class absence} (\mbox{ImageNet-C}, five passes, two runs).
Only the same 50 classes appear; the remaining 950 are absent.
Error (\%), mean over runs; SCALAR-C in gray (oracle reference). Bold: best non-oracle result.}
\label{tab:true_subset}
\footnotesize
\setlength{\tabcolsep}{3pt}
\begin{tabular}{lcccc}
\toprule
Method & Pass 1 & Pass 5 & Final clean & Coll. \\
\midrule
\multicolumn{5}{l}{\emph{Conservative}} \\
EATA & \textbf{51.2} & \textbf{44.2} & 18.3 & 0/2 \\
\rowcolor{orange!15}
\,+SCALAR-U & 51.6 & 48.7 & \textbf{17.6} & 0/2 \\
\textcolor{gray}{\,+SCALAR-C} & \textcolor{gray}{51.8} & \textcolor{gray}{42.6} & \textcolor{gray}{17.4} & \textcolor{gray}{0/2} \\
\midrule
\multicolumn{5}{l}{\emph{Aggressive}} \\
EATA & 70.7 & 76.9 & 64.1 & 1/2 \\
\rowcolor{orange!15}
\,+SCALAR-U & \textbf{50.0} & \textbf{52.9} & \textbf{22.2} & \textbf{0/2} \\
\textcolor{gray}{\,+SCALAR-C} & \textcolor{gray}{54.6} & \textcolor{gray}{37.3} & \textcolor{gray}{20.6} & \textcolor{gray}{0/2} \\
\bottomrule
\end{tabular}
\end{minipage}
\end{table}

\begin{table}[t]
\centering
\caption{
\textbf{Transfer to TENT} (ResNet-50, standard ImageNet-C stream, severity 5).
Corruption / post-stream clean error (\%), mean\,$\pm$\,SD over three random stream orderings, paired across methods.
At medium and high speeds, TENT collapses in 3/3 runs; all TENT+SCALAR runs remain stable.
}
\label{tab:backbone_generality}
\providecommand{\sd}[1]{\,{\scriptsize$\pm$#1}}
\setlength{\tabcolsep}{3.9pt}
\renewcommand{\arraystretch}{1.10}
\small

\begin{tabular}{lcc|cc|cc}
\toprule
& \multicolumn{2}{c|}{Optimal speed}
& \multicolumn{2}{c|}{Medium speed}
& \multicolumn{2}{c}{High speed} \\
Method & Avg Err & Clean Err & Avg Err & Clean Err & Avg Err & Clean Err \\
\midrule
TENT-R50~\cite{Tent2021}
& 67.5\sd{0.4} & 32.9\sd{0.3}
& 88.3\sd{1.3} &  97.1\sd{1.1}
& 95.9\sd{0.4} & 99.6\sd{0.2} \\
\rowcolor{orange!15}
\textbf{TENT + SCALAR}
&  66.5\sd{0.3} &  30.9\sd{0.2}
& \textbf{67.4}\sd{0.5} & \textbf{37.1}\sd{0.8}
& \textbf{69.8}\sd{1.1} & \textbf{43.3}\sd{1.2} \\
\addlinespace[0.2em]
\bottomrule
\end{tabular}
\end{table}

\subsection{Transfer Beyond ViT Teacher--Student Adaptation}
\label{sec:q3}

We also add SCALAR to TENT with a ResNet-50 backbone (Tab.~\ref{tab:backbone_generality}). This tests whether balancing helps a single model trained by entropy minimization, without an EMA teacher. SCALAR gives a small improvement at the stable operating point and larger gains when faster adaptation causes drift and forgetting. At high adaptation speed, where TENT collapses in all three runs, SCALAR reduces corruption error from $95.9\%$ to $69.8\%$ and clean error from $99.6\%$ to $43.3\%$.

\subsection{Ablation: Temporal Window and Hyperparameters}
\label{sec:ablation}

We examine the effect of the temporal window size $N$, temperature $\tau$, loss weight $\lambda$, and the number of Sinkhorn iterations $K$ on ImageNet-C. A longer window includes more predictions when balancing class usage, but also retains older predictions for longer.

We evaluate $N\in\{256,512,1024,4096\}$, $\tau\in[0.01,1.0]$, and $\lambda\in[0.1,0.9]$. Corruption error is $36.1\%$ at both $N=1024$ and $4096$, compared with $38.2\%$ at $N=512$. It varies by $1.1$ percentage points over $\tau\in[0.01,1.0]$ and by $0.5$ percentage points over $\lambda\in[0.3,0.7]$. Additional Sinkhorn iterations beyond $K=5$ did not yield measurable improvements. Detailed sensitivity curves are provided in App.~\ref{app:hyperparameters}.

\section{Conclusion}
\label{sec:conclusion}

We proposed SCALAR to control the class marginals of the pseudo-labels used for continual test-time adaptation. SCALAR balances recent predictions over a short window before they are used to update the model. Its benefit is largest in our experiments when class usage contracts and adaptation becomes unstable. It also reduces drift and collapse when added to EATA and TENT, and improves calibration under aggressive adaptation.

A uniform target provides stabilization without knowledge of the test-time prior. The correct prior can improve accuracy further, but the choice of target also matters when adaptation is already stable. Under persistent class absence, uniform balancing can limit useful specialization to the classes that remain.

\paragraph{Limitations and future work.}
Our experiments focus on image classification and controlled skewed streams constructed from corruption benchmarks. Further work is needed to test SCALAR on naturally occurring non-i.i.d.\ streams and on tasks such as detection and segmentation. A further question is how to adapt the target prior and balancing strength so that they allow specialization while still preventing collapse. 

\paragraph{Broader impact.}
Reducing adaptation-induced errors could improve the reliability of deployed image classifiers under changing conditions. However, forcing a uniform target can worsen performance under genuine class imbalance, so deployment requires validation on the intended population and monitoring for harmful adaptation.

\begin{ack}
Ofir Lindenbaum was supported by the MOST grant No. 0007341. The authors declare no competing interests.
\end{ack}
\bibliographystyle{plainnat}
\bibliography{SCALAR_arxiv_references}

\clearpage
\appendix

This appendix gives implementation details, hyperparameter analyses, per-corruption results, additional ablations, computational costs, and the construction of skewed streams.

\section{Implementation Details}
\label{app:implementation}
\subsection{Experimental Setup}

\noindent\textbf{Baselines and reproduction.}
For methods whose results are taken from prior CTTA papers, we use the reported numbers under the corresponding benchmark setting. As a check on our evaluation, we reproduce the source-only baseline using the ViT-B/16 checkpoint from our experiments. The reproduced source-only errors differ from the published values by at most 0.6 percentage points. Table~\ref{tab:reproduction} lists these reproduction checks. Our main controlled comparisons use paired runs with and without SCALAR under the same initialization.

\paragraph{Hyperparameters (balanced benchmarks).}
For CIFAR-10-C experiments, we use a batch size of 4, dual-rate EMA parameters $(\alpha_{\text{att}},\alpha_{\text{rest}})
= (0.90,0.99)$, and learning rate $\eta=5 \times 10^{-5}$.
CIFAR-100-C experiments use batch size 4, learning rate $\eta=5\times10^{-5}$,
and dual-rate EMA parameters
$(\alpha_{\text{att}},\alpha_{\text{rest}})=(0.9999,0.999)$.
ImageNet-C experiments use batch size 8, learning rate $\eta=10^{-5}$, and
dual-rate EMA parameters $(\alpha_{\text{att}},\alpha_{\text{rest}})=(0.90,0.99)$.
We chose EMA rates for stable adaptation of the teacher--student baseline without SCALAR and fixed them for each dataset. The ablation in App.~\ref{app:dual_vs_single_ema} compares single- and dual-rate EMA settings with SCALAR; it does not compare EMA sensitivity with and without SCALAR.

\subsection{Augmentation Pipeline}
\label{app:augmentations}
The teacher processes the incoming images without further augmentation, while the student receives strongly augmented views for the consistency objective.
Table~\ref{tab:aug} lists the student augmentations.
We use transformations intended to preserve the class label while changing appearance, such as color changes and small geometric distortions. These augmentations are used for the adaptation loss; predictions are made on inputs without these additional transformations.

\vspace{-2pt}
\begin{table}[t]
\centering
\caption{Reproduction checks against published benchmark
results. All values are average top-1 corruption error (\%).
Published results are retained in Table~\ref{tab:ctta_iid_all}.}
\label{tab:reproduction}
\begin{tabular}{lcc}
\toprule
Method / Dataset & Reported & Reproduced \\
\midrule
Source-only / CIFAR-10-C  & 28.1 & 27.9 \\
Source-only / CIFAR-100-C & 35.4 & 36.0 \\
Source-only / ImageNet-C  & 55.8 & 56.2 \\
VIDA / ImageNet-C         & 43.4 & 43.35 \\
TENT / ImageNet-C         & 51.0 & 50.48 \\
\bottomrule
\end{tabular}
\end{table}
\subsection{Experiment-Specific Settings}
\label{app:exp_settings}
Unless otherwise stated, all experiments use the SCALAR settings of Sec.~\ref{sec:setup} and the batch sizes of App.~A.1, and collapse is defined as in App.~\ref{app:skew}. The adaptation regimes in the main paper differ mainly in learning rate and, for the teacher--student pipeline, EMA momentum; we list these values below for each experiment.
\paragraph{Fixed-skew adaptation speeds (Tab.~\ref{tab:imagenet_skewed_speeds_compact_clean}).}
High speed uses $\eta=10^{-5}$ with $\alpha_{\text{att}}=\alpha_{\text{rest}}=0.9$; medium speed uses $\eta=5{\times}10^{-6}$ with $\alpha_{\text{att}}=\alpha_{\text{rest}}=0.99$; and low speed uses $\eta=5{\times}10^{-7}$ with $\alpha_{\text{att}}=\alpha_{\text{rest}}=0.999$.
\paragraph{EATA under dynamic skew (Tab.~\ref{tab:short_eata_scalar_lr}).}
When SCALAR is added to EATA, the queue stores the logits of the samples that EATA selects for the update, and the loss combines EATA's weighted entropy with the cross-entropy to the balanced targets, with weights $1-\lambda$ and $\lambda$. The queue retains the logits of the latest $N$ selected samples; steps with no selected samples skip the update and leave the queue unchanged. In all EATA experiments, with or without SCALAR, the Fisher regularizer is disabled. The stable operating point uses $\eta=10^{-5}$, chosen as the learning rate at which EATA alone remains stable across runs; the aggressive regime uses $\eta=6{\times}10^{-4}$. EATA uses batch size 8, entropy margin $E_0=\ln(1000)/2-1$, and redundancy margin $\epsilon=0.05$.

\paragraph{Long-horizon experiments (Tabs.~\ref{tab:long_horizon_persistent_skew}--\ref{tab:true_subset}).}
The persistent-skew stream uses $\eta=3{\times}10^{-4}$. On the class-absence stream, the conservative regime uses $\eta=5{\times}10^{-5}$ and the aggressive regime uses $\eta=3{\times}10^{-4}$.

\paragraph{TENT (Tab.~\ref{tab:backbone_generality}).}
Since TENT has no teacher, SCALAR stores the model's own logits in the queue, and the loss combines the entropy with the cross-entropy to the balanced targets, with weights $1-\lambda$ and $\lambda$. TENT uses batch size 64; optimal, medium, and high speeds use $\eta=2.5{\times}10^{-4}$, $8{\times}10^{-2}$, and $1.5{\times}10^{-1}$, respectively.

\paragraph{Temporal contraction (Fig.~\ref{fig:dynamics_across_regimes}).}
The balanced stream uses the ImageNet-C settings of App.~\ref{app:implementation}; the fixed- and dynamic-skew streams use the medium and high speeds of Tab.~\ref{tab:imagenet_skewed_speeds_compact_clean}, respectively.

\begin{table}[ht]
\centering
\scriptsize
\caption{Student augmentation pipeline for consistency training. Strong augmentations are applied only to student inputs; teacher inputs are not augmented.}
\label{tab:aug}

\begin{tabular}{lp{5.25cm}}
\toprule
\textbf{Augmentation} & \textbf{Parameters + note} \\ \midrule
RandomAffine &
\shortstack[l]{degrees $0^{\circ}$\\
translate $(32/W,\;32/H)$\\
scale $0.98$–$1.02$,\; bilinear,\; fill=0
} \\[4pt]

ColorJitterPro &
\shortstack[l]{brightness $0.6$–$1.4$, contrast $0.7$–$1.3$,\\
saturation $0.5$–$1.5$, hue $\pm0.06$,\\
gamma $0.7$–$1.3$} \\[4pt]

RandomGrayscale & $p = 0.2$ \\ \bottomrule
\end{tabular}
\end{table}

\section{Additional Details on Sinkhorn Balancing}
\label{app:sinkhorn}

\subsection{KL-Projection Interpretation}
\label{sec:kl}

The balancing step seeks targets that satisfy the desired class marginals while staying close to the teacher predictions in KL divergence~\cite{sinkhorn1964,cuturi2013}. The constraints apply to the whole window: individual samples can still have different class distributions. Below, we give the window construction and normalization steps used to approximate this projection.

\paragraph{Window of teacher predictions.}
Let $B$ be the current batch size and $M = N - B$ the number of past logits retained in the queue, so that the window has $N$ rows as in Sec.~\ref{sec:method}. 
We stack the teacher logits from the current batch and the queue:
\[
\text{logits}_{\text{eff}} \in \mathbb{R}^{(B+M)\times C}.
\]
We convert logits to probabilities with a temperature $\tau>0$:
\[
P = \mathrm{softmax}\big(\text{logits}_{\text{eff}}/\tau\big) \in \mathbb{R}^{N\times C}, 
\quad N = B+M.
\]
Each row of $P$ is a class distribution for one sample. The column sums give the total predicted mass per class; dividing them by $N$ gives the average distribution over the window.

\paragraph{Sinkhorn balancing to a prior.}
We seek $Q \in \mathbb{R}^{N\times C}$ with row sums~1 and column
marginals $N\pi \in \mathbb{R}^C$:
\[
Q\mathbf{1} = \mathbf{1}, 
\qquad 
\mathbf{1}^\top Q = N\,\pi^\top,
\]
where $\pi \in \Delta^{C-1}$ encodes the desired class usage
(uniform $\pi_c = 1/C$ when no test-time prior is known, or a given
prior when class skew is specified). We obtain $Q$ by applying $K$ rounds of Sinkhorn scaling.
Starting from $Q^{(0)} = P$, each Sinkhorn iteration consists of:
\[
\tilde{Q}^{(k)} \gets Q^{(k)} \oslash \big(Q^{(k)}\mathbf{1}\big),
\]
\[
Q^{(k+1)} \gets \tilde{Q}^{(k)}\,\mathrm{Diag}\!\Big(\dfrac{N\,\pi}{\tilde{Q}^{(k)\top}\mathbf{1}_N}\Big),
\qquad k = 0,\dots,K-1,
\]
and a final row normalization
\[
Q \gets Q^{(K)} \oslash \big(Q^{(K)}\mathbf{1}\big).
\]
Here $\oslash$ denotes row-wise division by the current row sums
and $\mathrm{Diag}(\cdot)$ forms a diagonal matrix, as in Eq.~\eqref{eq:sinkhorn}. The final normalization ensures that each row sums to one. Since we use a finite number of iterations, the column sums need only approximately match $N\pi$; $Q$ therefore approximates the KL projection onto the specified marginals. In the implementation, target column masses are clamped to at least $10^{-12}$ and $10^{-12}$ is added to all row and column sums, which avoids division by zero for classes with zero prior mass.

\paragraph{Objective and update.}
Let $q_u$ denote the original teacher probabilities and $\tilde q_u$ the balanced targets for the current batch obtained by Sinkhorn projection toward the target prior $\pi$. The SCALAR-specific supervision term is
\[
\mathcal{L}_{\mathrm{bal}} = \mathrm{CE}(\mathrm{sg}[\tilde q_u],\, p_v).
\]
The combined objective is
\[
\mathcal{L} = (1-\lambda)\,\mathcal{L}_{\mathrm{consistency}} + \lambda\,\mathcal{L}_{\mathrm{bal}},
\]
which is identical to Eq.~\eqref{eq:total-loss}. During warm-up, before marginal balancing is activated, the method reduces to $\mathcal{L}_{\mathrm{consistency}}$.

We apply a single gradient step on the student parameters $\theta$ and then update the teacher via the dual-rate EMA. SCALAR modifies the teacher targets used in the loss without changing the backbone architecture.

\begin{figure*}[t]
\centering
\begin{minipage}{0.32\textwidth}
\centering
\begin{tikzpicture}
\begin{axis}[
  width=\linewidth, height=0.9\linewidth,
  xlabel={Window size}, ylabel={Error (\%)},
  xmode=log, log basis x=2,
  xtick={256,512,1024,4096},
  xticklabels={256,512,1024,4096},
  log ticks with fixed point,
  scaled x ticks=false,
  xticklabel style={font=\footnotesize, yshift=-1pt},
  ymin=35.5, ymax=38.5,
  enlarge x limits=0.08, yminorticks=false
]
\addplot+[mark=*] coordinates {(256,38.1) (512,38.2) (1024,36.1) (4096,36.1)};
\end{axis}
\end{tikzpicture}
\end{minipage}\hfill
\begin{minipage}{0.32\textwidth}
\centering
\begin{tikzpicture}
\begin{axis}[
  width=\linewidth, height=0.9\linewidth,
  xlabel={Temperature $\tau$}, ylabel={Error (\%)},
  xmode=log, log basis x=10,
  xtick={0.01,0.1,0.3,0.7,1.0},
  xticklabels={0.01,0.1,0.3,0.7,1.0},
  x tick label style={rotate=90,anchor=east,font=\scriptsize},
  log ticks with fixed point,
  scaled x ticks=false,
  xticklabel style={font=\footnotesize, yshift=-1pt},
  ymin=35.5, ymax=37.5,
  enlarge x limits=0.08, yminorticks=false
]
\addplot+[mark=*] coordinates {(0.01,36.8) (0.1,36.1) (0.3,36.3) (0.7,36.8) (1.0,37.2)};
\end{axis}
\end{tikzpicture}
\end{minipage}\hfill
\begin{minipage}{0.32\textwidth}
\centering
\begin{tikzpicture}
\begin{axis}[
  width=\linewidth, height=0.9\linewidth,
  xlabel={Loss weight $\lambda$}, ylabel={Error (\%)},
  xtick={0.1,0.3,0.5,0.7,0.9},
  xticklabels={0.1,0.3,0.5,0.7,0.9},
  scaled x ticks=false,
  xticklabel style={font=\footnotesize, yshift=-1pt},
  ymin=35.5, ymax=37.5,
  enlarge x limits=0.05, yminorticks=false
]
\addplot+[mark=*] coordinates {(0.1,37.1) (0.3,36.6) (0.5,36.1) (0.7,36.2) (0.9,36.9)};
\end{axis}
\end{tikzpicture}
\end{minipage}
\caption{\textbf{SCALAR hyperparameter sensitivity on ImageNet-C (class-balanced corruption setting, severity 5; error \%).}}
\label{fig:scalar_ablation}
\end{figure*}
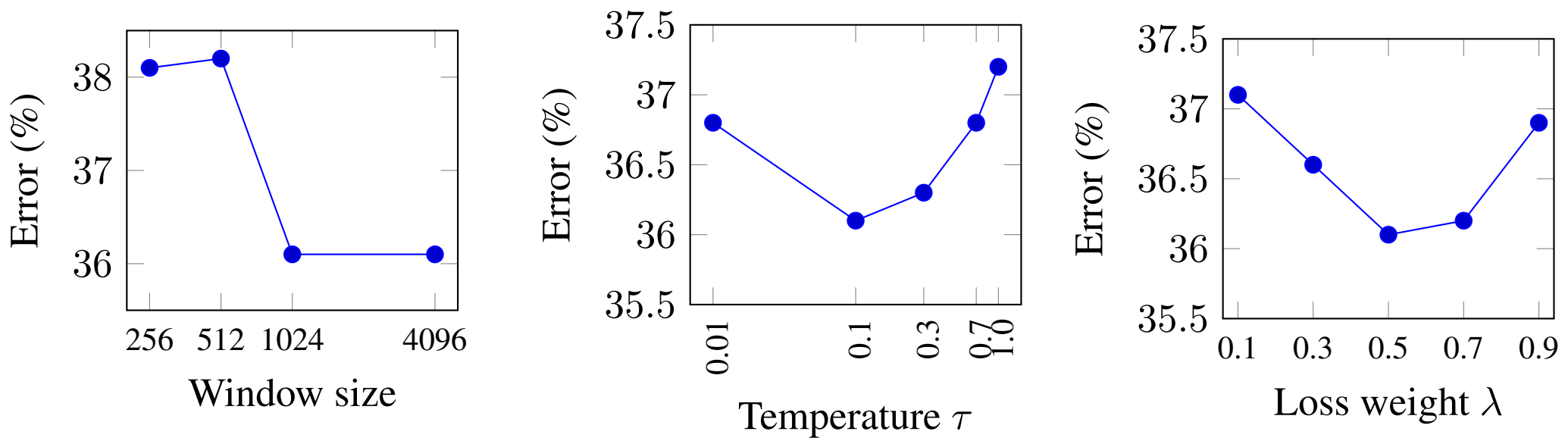

\section{Hyperparameter Sensitivity}
\label{app:hyperparameters}

\subsection{SCALAR module ablations (queue + Sinkhorn)}
\label{sec:ablations-module}

We vary the window size, temperature, and loss weight to examine how each affects SCALAR. All experiments in this section use a uniform target and $K {=} 5$ Sinkhorn iterations.

Figure~\ref{fig:scalar_ablation} summarizes the sensitivity of SCALAR to its three
primary hyperparameters on ImageNet-C in the standard corruption setting:
(i) the mixing weight $\lambda$ controlling the strength of the SCALAR term,
(ii) the temperature $\tau$ applied in the softmax over the queue window, and
(iii) the window size $N$ (queue length).

Windows of 256--512 samples give higher error than windows of 1024--4096 samples. Increasing the window from $N=1024$ to $4096$ gives no further reduction in this sweep. Thus, using more recent predictions helps up to a point, but the largest tested window offers no advantage over our default.

Error is lowest at $\tau=0.1$ and $\lambda=0.5$, with slightly higher values at either end of the tested ranges. The mixing-weight sweep under class skew (App.~\ref{app:lambda_skew}) also gives lower corruption errors at $\lambda=0.5$--$0.75$ than at $0.25$.

\section{Per-Corruption Results}
\label{app:corruptions}
The following tables report results for all 15 corruption types under the default configurations used in the main paper.

In Tables~\ref{tab:cifar10c_scalar}--\ref{tab:imagenetc_scalar}, boldface
indicates the best method for each corruption.
On CIFAR-10-C, SCALAR attains the lowest error for 13 of 15 corruptions, and on ImageNet-C, it is best on 13 of 15 as well, while also achieving the lowest average error and the largest gain.
On CIFAR-100-C, SCALAR is best on 4 out of 15 corruptions, but still attains the best average error and gain.

\begin{table*}[t]
\centering
\scriptsize
\setlength{\tabcolsep}{3pt}

\caption{\textbf{CIFAR-10-C (severity 5) corruption results (error \%, lower is better).}
Per-corruption test error and average over all 15 corruptions.
\emph{Gain} is the improvement in average error relative to the original
Source model, in percentage points (higher is better). \textbf{Boldface} indicates the best method for each
corruption. SCALAR results are averaged over 5 seeds; the standard deviation of its average error is below 0.2 percentage points. Prior-method values are taken from the cited publications.}
\label{tab:cifar10c_scalar}

\resizebox{\textwidth}{!}{%
\begin{tabular}{l l *{15}{c} c c}
\toprule
Time & &
\multicolumn{15}{c}{$t \xrightarrow{\hspace*{12.2cm}}$} & & \\
Method & Ref &
Gauss. & Shot & Impulse & Defocus & Glass & Motion & Zoom &
Snow & Frost & Fog & Bright. & Contrast & Elastic & Pixelate & JPEG &
Avg. & Gain \\
\midrule
Source &  &
60.1 & 53.2 & 38.3 & 19.9 & 35.5 & 22.6 & 18.6 & 12.1 & 12.7 & 22.8 &
5.3 & 49.7 & 23.6 & 24.7 & 23.1 &
28.1 & 0.0 \\[0.15em]
TENT & ~\cite{Tent2021} &
57.7 & 56.3 & 29.4 & 16.2 & 35.3 & 16.2 & 12.4 & 11.0 & 11.6 & 14.9 &
4.7 & 22.5 & 15.9 & 29.1 & 19.5 &
23.5 & 4.6 \\[0.15em]
CoTTA & ~\cite{CoTTA2022} &
58.7 & 51.3 & 33.0 & 20.1 & 34.8 & 20.0 & 15.2 & 11.1 & 11.3 & 18.5 &
4.0 & 34.7 & 18.8 & 19.0 & 17.9 &
24.6 & 3.5 \\[0.15em]
VDP & ~\cite{VDP2023} &
57.5 & 49.5 & 31.7 & 21.3 & 35.1 & 19.6 & 15.1 & 10.8 & 10.3 & 18.1 &
4.0 & 27.5 & 18.4 & 22.5 & 19.9 &
24.1 & 4.0 \\[0.15em]
VIDA & ~\cite{VIDA2024} &
52.9 & 47.9 & 19.4 & 11.4 & 31.3 & 13.3 & 7.6 & 7.6 & 9.9 & 12.5 &
3.8 & 26.3 & 14.4 & 33.9 & 18.2 &
20.7 & 7.4 \\[0.15em]
Versatile & ~\cite{versatile2024} & \textbf{16.3} & 11.1 & 9.6 & 8.4 & 14.6 & 8.6 & 5.5 & 6.3 & 5.7 & 7.1 &
\textbf{3.3} & 5.4 & 10.9 & 7.7 & 12.8 &
8.9 & 19.2 \\[0.15em]
REM & ~\cite{REM2025} & 17.3 & 12.5 & 10.3 & 8.4 & 17.7 & 8.4 & 5.5 & 6.6 & 5.6 & 7.2 &
3.7 & 6.4 & 11.0 & 7.3 & 13.0 &
9.4 & 18.7 \\[0.15em]
\rowcolor{orange!15}
\textbf{SCALAR} & \textbf{Proposed} &
20.32 & \textbf{10.32} & \textbf{9.52} & \textbf{7.68} &
\textbf{12.14} & \textbf{7.76} & \textbf{5.42} & \textbf{5.94} &
\textbf{5.06} & \textbf{6.12} & 3.48 & \textbf{3.82} &
\textbf{8.64} & \textbf{5.58} & \textbf{9.62} &
\textbf{8.1} & \textbf{20.0} \\
\bottomrule
\end{tabular}%
}

\end{table*}

\begin{table*}[t]
\centering
\scriptsize
\setlength{\tabcolsep}{3pt}

\caption{\textbf{CIFAR-100-C (severity 5) corruption results (error \%, lower is better).}
Per-corruption test error and average over all 15 corruptions.
\emph{Gain} is the improvement in average error relative to the original Source model, in percentage points (higher is better). \textbf{Boldface} indicates the best method for each corruption. SCALAR results are averaged over 5 seeds; the standard deviation of its average error is below 0.2 percentage points. Prior-method values are taken from the cited publications.}
\label{tab:cifar100c_scalar}

\resizebox{\textwidth}{!}{%
\begin{tabular}{l l *{15}{c} c c}
\toprule
Time & &
\multicolumn{15}{c}{$t \xrightarrow{\hspace*{12.2cm}}$} & & \\
Method & Ref &
Gauss. & Shot & Impulse & Defocus & Glass & Motion & Zoom &
Snow & Frost & Fog & Bright. & Contrast & Elastic & Pixelate & JPEG &
Avg. & Gain \\
\midrule
Source        &   & 55.0  & 51.5  & 26.9  & 24.0  & 60.5  & 29.0  & 21.4  & 21.1  & 25.0  & 35.2  & 11.8  & 34.8  & 43.2  & 56.0  & 35.9  & 35.4  & 0   \\
TENT          & ~\cite{Tent2021}  & 53.0  & 47.0  & 24.6  & 22.3  & 58.5  & 26.5  & 19.0  & 21.0  & 23.0  & 30.1  & 11.8  & 25.2  & 39.0  & 47.1  & 33.3  & 32.1  & 3.3 \\
CoTTA         & ~\cite{CoTTA2022}  & 55.0  & 51.3  & 25.8  & 24.1  & 59.2  & 28.9  & 21.4  & 21.0  & 24.7  & 34.9  & 11.7  & 31.7  & 40.4  & 55.7  & 35.6  & 34.8  & 0.6 \\
VDP           & ~\cite{VDP2023}  & 54.8  & 51.2  & 25.6  & 24.2  & 59.1  & 28.8  & 21.2  & 20.5  & 23.3  & 33.8  &  \textbf{7.5}  & \textbf{11.7}  & 32.0  & 51.7  & 35.2  & 32.0  & 3.4 \\
VIDA          & ~\cite{VIDA2024}  & 50.1  & 40.7  & 22.0  & 21.2  & 45.2  & 21.6  & \textbf{16.5}  & 17.9  & 16.6  & 25.6  & 11.5  & 29.0  & 29.6  & 34.7  & 27.1  & 27.3  & 8.1 \\
Versatile     & ~\cite{versatile2024}          & 38.2  & 31.8  & 18.2  & 20.8  & 34.3  & \textbf{20.3}  & 17.5  & \textbf{14.9}  & \textbf{16.2}  & 22.9  & 11.5  & 27.5  & 28.2  & 32.5  & \textbf{25.3}  & 24.0  & 11.4 \\
ADMA          & ~\cite{ADMA2024}  & 48.6  & 30.7  & 18.5  & 21.3  & 38.4  & 22.2  & 17.5  & 19.3  & 18.0  & 24.8  & 13.1  & 27.8  & 31.4  & 35.5  & 29.5  & 26.4  & 9.0 \\
REM           & ~\cite{REM2025}          & \textbf{29.2}  & \textbf{25.5}  & \textbf{17.0}  & \textbf{19.1}  & 35.2  & 21.2  & 18.3  & 19.5  & 18.7  & 22.8  & 15.5  & 17.6  & 31.6  & 26.2  & 33.0  & 23.4  & 12.0 \\
\rowcolor{orange!15}
\textbf{SCALAR}        & \textbf{Proposed}          & 38.26 & 26.98 & 21.6  & 23.5  & \textbf{28.78}& 22.66 & 20.16 & 20.48 & 19.02 & \textbf{21.2}  & 16.5  & 16.82 & \textbf{25.6}  & \textbf{21.06} & 25.44 & \textbf{23.2}  & \textbf{12.2} \\
\bottomrule
\end{tabular}%
}
\end{table*}

\begin{table*}[t]
\centering
\scriptsize
\setlength{\tabcolsep}{3pt}

\caption{\textbf{ImageNet-C (severity 5) corruption results (error \%, lower is better).}
Per-corruption test error and average over all 15 corruptions.
\emph{Gain} is the improvement in average error relative to the original
Source model, in percentage points (higher is better). \textbf{Boldface} indicates the best method for each
corruption. SCALAR results are averaged over 5 seeds; the standard deviation of its average error is below 0.2 percentage points. Prior-method values are taken from the cited publications. Versatile's reported mean of $42.7\%$ is retained; its displayed entries average to $42.77\%$.}
\label{tab:imagenetc_scalar}

\resizebox{\textwidth}{!}{%
\begin{tabular}{l l *{15}{c} c c}
\toprule
Time & &
\multicolumn{15}{c}{$t \xrightarrow{\hspace*{12.2cm}}$} & & \\
Method & Ref &
Gauss. & Shot & Impulse & Defocus & Glass & Motion & Zoom &
Snow & Frost & Fog & Bright. & Contrast & Elastic & Pixelate & JPEG &
Avg. & Gain \\
\midrule
Source        &   & 53.0  & 51.8  & 52.1  & 68.5  & 78.8  & 58.5  & 63.3  & 49.9  & 54.2  & 57.7  & 26.4  & 91.4  & 57.5  & 38.0  & 36.2  & 55.8  & 0.0  \\
TENT          & ~\cite{Tent2021}  & 52.2  & 48.9  & 49.2  & 65.8  & 73.0  & 54.5  & 58.4  & 44.0  & 47.7  & 50.3  & 23.9  & 72.8  & 55.7  & 34.4  & 33.9  & 51.0  & 4.8 \\
CoTTA         &  ~\cite{CoTTA2022}  & 52.9  & 51.6  & 51.4  & 68.3  & 78.1  & 57.1  & 62.0  & 48.2  & 52.7  & 55.3  & 25.9  & 90.0  & 56.4  & 36.4  & 35.2  & 54.8  & 1.0 \\
VDP           &  ~\cite{VDP2023}  & 52.7  & 51.6  & 50.1  & 58.1  & 70.2  & 56.1  & 58.1  & 42.1  & 46.1  & 45.8  & 23.6  & 70.4  & 54.9  & 34.5  & 36.1  & 50.0  & 5.8 \\
VIDA          &  ~\cite{VIDA2024}  & 47.7  & 42.5  & 42.9  & 52.2  & 56.9  & 45.5  & 48.9  & 38.9  & 42.7  & 40.7  & 24.3  & 52.8  & 49.1  & 33.5  & 33.1  & 43.4  & 12.4 \\
Versatile     & ~\cite{versatile2024} & 47.5  & 42.1  & 41.6  & 55.5  & 55.4  & 44.5  & 47.9  & 38.8  & 37.8  & 39.6  & 23.6  & 57.0  & 44.4  & 33.5  & 32.3  & 42.7  & 13.1 \\
ADMA          &  ~\cite{ADMA2024}  & 46.3  & 41.9  & 42.5  & 51.4  & 54.9  & 43.3  & 40.7  & 34.2  & 35.8  & 64.3  & 23.4  & 60.3  & 37.5  & 29.2  & 31.4  & 42.5  & 13.3 \\
REM           &   ~\cite{REM2025}         & \textbf{43.5}  & \textbf{38.1}  & 39.2  & 53.2  & 49.0  & 43.5  & 42.8  & 37.5  & 35.2  & 35.4  & 23.2  & 46.8  & 41.6  & 28.9  & 30.2  & 39.2  & 16.6 \\
\rowcolor{orange!15}
SCALAR        &  \textbf{Proposed}         & 46.4  & 39.78 & \textbf{38.88} & \textbf{46.24} & \textbf{44.28} & \textbf{38.2}  & \textbf{37.02} & \textbf{33.28} & \textbf{33.22} & \textbf{32.42} & \textbf{21.74} & \textbf{40.9}  & \textbf{34.36} & \textbf{26.76} & \textbf{28.54} & \textbf{36.1}  & \textbf{19.7} \\
\bottomrule
\end{tabular}
}

\end{table*}

\section{Additional Ablation Studies}
\label{app:ablations}
\subsection{Dual vs.\ Single EMA on CIFAR-100-C}
\label{app:dual_vs_single_ema}
\paragraph{Dual-rate EMA teacher.}
We use an EMA teacher with separate decay rates for attention blocks and the remaining layers. 
Let $\theta^{\text{att}}$ and $\theta^{\text{rest}}$ denote the student parameters for attention and for the rest of the network (MLP, convolution, embeddings), and $\phi^{\text{att}}, \phi^{\text{rest}}$ the corresponding teacher parameters. 
After each student update, we apply
\begin{align*}
\phi^{\text{att}}  &\leftarrow \alpha_{\text{att}}  \,\phi^{\text{att}} 
                  + (1 - \alpha_{\text{att}})\,\theta^{\text{att}}, \\
\phi^{\text{rest}} &\leftarrow \alpha_{\text{rest}} \,\phi^{\text{rest}} 
                  + (1 - \alpha_{\text{rest}})\,\theta^{\text{rest}},
\end{align*}
with $(\alpha_{\text{att}}, \alpha_{\text{rest}})$ fixed by the teacher--student baseline configuration, independently of SCALAR.
Separate rates allow attention and the remaining parameters to follow student updates at different speeds. We use dual-rate EMA in our ViT teacher--student experiments; SCALAR can also be used with other teacher--student designs or without an EMA teacher.

To study the effect of using separate EMA rates for different layers, we run a
controlled ablation on CIFAR-100-C (severity 5) with SCALAR.
We fix the learning rate and SCALAR hyperparameters (uniform prior, temperature
$\tau = 0.1$, window size $N = 1024$, Sinkhorn iterations $K = 5$) and vary
only the EMA configuration.

Tab.~\ref{tab:cifar100_ema_full} compares a \emph{single-EMA} scheme, where a
shared rate $\alpha$ is used for all parameters, and a \emph{dual-EMA} scheme,
where attention blocks and the remaining weights use different EMA rates
$(\alpha_{\text{rest}}, \alpha_{\text{att}})$.
For the single-EMA setting we sweep $\alpha$ in the range
$[0.9990, 0.9999]$ in steps of $10^{-4}$ (with an additional point at
$\alpha = 0.99985$), while the dual-EMA configuration uses
$(\alpha_{\text{rest}}, \alpha_{\text{att}}) = (0.999, 0.9999)$.

The average corruption error is measured on the CIFAR-100-C severity-5 stream, and the clean-post error is measured on CIFAR-100 after completing the stream.
Within this sweep, the dual-EMA configuration achieves the lowest observed corruption error ($22.9\%$) and reduces the clean-post error below the initial clean error (from $\approx 17.3\%$ to $16.7\%$).
The slowest single-EMA setting ($\alpha = 0.9999$) achieves the best clean-post error in this grid ($15.9\%$), but at the cost of a higher average corruption
error ($25.6\%$), roughly $2$--$3$ percentage points above the dual-EMA value.
Even when sampling single-EMA rates as fine as $\alpha = 0.99985$, we do not find a single rate that matches the dual-EMA corruption–clean trade-off in this configuration.
Separate EMA rates help in this CIFAR-100-C experiment. More datasets and hyperparameter settings are needed to test whether the benefit holds more generally.

\begin{table*}[t]
\centering
\scriptsize
\setlength{\tabcolsep}{3pt}

\caption{\textbf{Single vs.\ dual EMA on CIFAR-100-C (severity 5).} Each row reports per-corruption error (\%) on the CIFAR-100-C severity-5 stream, along with the average corruption error (\emph{Avg.}) and the post-stream clean error on CIFAR-100 (\emph{Clean}).
Single-EMA rows use a shared rate for all parameters ($\alpha_{\text{att}} = \alpha_{\text{rest}}$), whereas the dual-EMA configuration decouples attention and the remaining weights.}
\label{tab:cifar100_ema_full}

\resizebox{\textwidth}{!}{
\begin{tabular}{c cc *{15}{c} c c}
\toprule
LR & $\alpha_{\text{rest}}$ & $\alpha_{\text{att}}$ &
Gauss. & Shot & Impulse & Defocus & Glass & Motion & Zoom &
Snow & Frost & Fog & Bright. & Contrast & Elastic & Pixelate & JPEG &
Avg. & Clean \\
\midrule
$5\times10^{-5}$ & 0.9990   & 0.9990   &
40.61 & 30.07 & 25.03 & 26.99 & 32.66 & 24.94 & 22.76 &
23.75 & 22.07 & 25.28 & 19.25 & 19.57 & 31.42 & 29.34 & 32.71 &
27.10 & 25.30 \\[0.15em]
$5\times10^{-5}$ & 0.9991   & 0.9991   &
40.00 & \textbf{28.84} & 24.43 & 26.45 & 33.75 & 25.66 & 23.52 &
25.63 & 23.19 & 26.36 & 21.08 & 20.98 & 32.70 & 29.66 & 32.57 &
27.70 & 26.66 \\[0.15em]
$5\times10^{-5}$ & 0.9992   & 0.9992   &
41.08 & 29.82 & 23.95 & 26.38 & 31.99 & 24.83 & 22.23 &
23.64 & 21.46 & 24.87 & 19.06 & 19.43 & 31.31 & 28.88 & 32.87 &
26.80 & 25.38 \\[0.15em]
$5\times10^{-5}$ & 0.9993   & 0.9993   &
41.84 & 30.24 & 25.31 & 27.02 & 34.87 & 25.77 & 22.24 &
24.21 & 22.60 & 27.14 & 19.88 & 19.97 & 31.95 & 28.05 & 31.20 &
27.50 & 24.66 \\[0.15em]
$5\times10^{-5}$ & 0.9994   & 0.9994   &
42.22 & 30.69 & 24.05 & 25.94 & 32.54 & 24.36 & 21.43 &
23.27 & 20.91 & 24.35 & 18.22 & 18.42 & 29.80 & 27.57 & 29.88 &
26.20 & 22.56 \\[0.15em]
$5\times10^{-5}$ & 0.9995   & 0.9995   &
43.48 & 31.34 & 24.91 & 27.29 & 33.43 & 24.87 & 21.60 &
23.17 & 20.85 & 24.69 & 18.13 & 18.71 & 30.59 & 31.42 & 29.85 &
27.00 & 22.63 \\[0.15em]
$5\times10^{-5}$ & 0.9996   & 0.9996   &
44.44 & 31.92 & 23.41 & 25.95 & 33.07 & 24.69 & 21.80 &
22.93 & 20.52 & 25.18 & 17.59 & 18.45 & 29.21 & 27.75 & 28.52 &
26.40 & 21.20 \\[0.15em]
$5\times10^{-5}$ & 0.9997   & 0.9997   &
47.44 & 36.02 & 23.51 & 25.30 & 33.43 & 23.65 & 19.61 &
20.66 & 18.99 & 23.34 & 15.56 & 16.65 & 27.82 & 29.03 & 27.84 &
25.90 & 19.49 \\[0.15em]
$5\times10^{-5}$ & 0.9998   & 0.9998   &
49.20 & 38.28 & 22.70 & 24.75 & 36.84 & 23.36 & 19.27 &
20.01 & 18.39 & 23.00 & 15.04 & 16.02 & 27.40 & 28.78 & 26.91 &
26.00 & 18.35 \\[0.15em]
$5\times10^{-5}$ & 0.99985  & 0.99985  &
50.17 & 40.31 & 22.52 & 24.94 & 36.12 & 23.69 & 18.94 &
19.26 & 17.72 & 22.66 & 13.70 & 15.84 & 26.41 & 28.27 & 25.67 &
25.70 & 17.14 \\[0.15em]
$5\times10^{-5}$ & 0.9999   & 0.9999   &
51.47 & 42.70 & 22.45 & 23.59 & 37.90 & 22.53 & 17.81 &
\textbf{17.73} & \textbf{16.36} & 21.80 & \textbf{12.34} & \textbf{14.78} & 24.71 & 31.79 & 25.71 &
25.60 & \textbf{15.92} \\[0.25em]
$5\times10^{-5}$ &
\textbf{0.9990} & \textbf{0.9999} &
\textbf{43.56} & 32.51 & \textbf{20.39} & \textbf{22.05} &
\textbf{31.25} & \textbf{20.69} & \textbf{17.53} &
18.32 & 16.67 & \textbf{20.55} & 13.66 &
15.43 & \textbf{24.12} & \textbf{23.76} & \textbf{23.64} &
\textbf{22.90} & 16.73 \\
\bottomrule
\end{tabular}
}
\end{table*}

\subsection{Batch-Size Robustness}
\label{app:batch_size}

We examine the effect of batch size on EATA+SCALAR-U
under persistent class skew over five consecutive passes (Tab.~\ref{tab:batch_size}).
The same 50 classes remain dominant throughout the stream.
We vary the batch size while keeping the window size fixed
at $N=1024$, with learning rate $\eta=10^{-5}$ and
temperature $\tau=0.3$. This is a separate configuration from the persistent-skew experiment in Tab.~\ref{tab:long_horizon_persistent_skew}, which uses $N=4096$, $\tau=0.7$, and $\eta=3\times10^{-4}$; the absolute errors are therefore not directly comparable.

\begin{table}[t]
\centering
\caption{Batch-size sensitivity on ImageNet-C (severity 5)
under five-pass persistent skew. Error (\%), mean $\pm$
standard deviation over three runs.}
\label{tab:batch_size}
\begin{tabular}{lccc}
\toprule
Batch size & Avg. error & Final clean & Coll. \\
\midrule
2 & $49.2 \pm 1.2$ & $20.3 \pm 0.3$ & 0/3 \\
4 & $47.7 \pm 0.6$ & $19.9 \pm 0.5$ & 0/3 \\
8 & $46.0 \pm 0.8$ & $19.3 \pm 0.1$ & 0/3 \\
\bottomrule
\end{tabular}
\end{table}

SCALAR remains stable across the tested batch sizes,
including batches of two samples. Larger batches show
modestly lower corruption and post-stream clean errors
in this setting.

\subsection{Sensitivity to the Mixing Weight under Class Skew}
\label{app:lambda_skew}

We also examine the sensitivity to the SCALAR
mixing weight $\lambda$ under three levels of class skew,
with 25, 50, and 200 dominant classes (Tab.~\ref{tab:lambda_skew}). We use a window
size of $N=1024$ and report a single run per configuration.

\begin{table}[t]
\centering
\caption{Sensitivity to the mixing weight under class skew
on ImageNet-C (severity 5). Each entry reports average
corruption error / post-stream clean error (\%).}
\label{tab:lambda_skew}
\begin{tabular}{lccc}
\toprule
Dominant classes & $\lambda=0.25$ & $\lambda=0.5$ & $\lambda=0.75$ \\
\midrule
25  & 50.7 / 21.6 & 43.9 / 18.5 & 45.8 / 18.0 \\
50  & 49.6 / 19.4 & 48.4 / 18.8 & 48.7 / 18.1 \\
200 & 48.9 / 20.3 & 46.7 / 19.2 & 46.6 / 18.3 \\
\midrule
Mean & 49.7 / 20.4 & 46.3 / 18.9 & 47.0 / 18.1 \\
\bottomrule
\end{tabular}
\end{table}

The choice of mixing weight affects corruption and clean error differently. Averaged across the tested skew levels, $\lambda=0.5$ gives the lowest corruption error, while $\lambda=0.75$ gives the lowest post-stream clean error. Each configuration uses one run, so these comparisons do not establish a precise optimum.

\subsection{Comparison with Target Sharpening}
\label{app:sharpening}

We compare SCALAR with a target-sharpening baseline in the
fixed-skew, high-speed ImageNet-C setting of
Tab.~\ref{tab:imagenet_skewed_speeds_compact_clean}.
The sharpened variant replaces the balanced targets
$\tilde{q}_u$ in Eq.~\eqref{eq:total-loss} with
$\mathrm{softmax}(z_u/\tau)$, keeping the same warm-up
length, temperature $\tau=0.1$, and mixing weight
$\lambda=0.5$. The sharpening baseline is evaluated over
three independent runs under matched adaptation conditions.
The results are shown in Tab.~\ref{tab:sharpening}.

\begin{table}[t]
\centering
\caption{Comparison with target sharpening on fixed-skew
ImageNet-C (severity 5, high speed). Average corruption error
and final clean error (\%), mean $\pm$ standard deviation
over three runs. Coll. indicates collapsed runs. Baseline
and SCALAR-U results are from
Tab.~\ref{tab:imagenet_skewed_speeds_compact_clean}.}
\label{tab:sharpening}
\begin{tabular}{lccc}
\toprule
Method & Avg. error & Final clean & Coll. \\
\midrule
Baseline (w/o SCALAR) & $86.8 \pm 14.4$ & $76.4 \pm 40.6$ & 2/3 \\
Sharpening            & $90.7 \pm 6.5$  & $74.7 \pm 43.3$ & 2/3 \\
SCALAR-U              & $48.1 \pm 1.1$  & $21.1 \pm 0.5$  & 0/3 \\
\bottomrule
\end{tabular}
\end{table}

Sharpening does not give a consistent improvement over the
baseline, with both methods collapsing in two of the three
runs. SCALAR-U gives lower corruption and final clean errors
than sharpening in this comparison, suggesting that
sharpening targets alone does not account for the benefit
of balancing across the window.

\section{Computational Overhead}
\label{app:overhead}
SCALAR converts an $N \times C$ FIFO queue of teacher logits to probabilities and applies $K$ iterations of Sinkhorn row/column scaling.
Ignoring the lower-order cost of the per-step softmax on the current mini-batch
(typically negligible when $N \gg B$), each Sinkhorn iteration consists of two
reductions and two elementwise normalization passes:
\emph{row normalization}
$r = Q \mathbf{1}_C$, $Q \leftarrow Q \oslash r$,
followed by
\emph{column scaling}
$c = \mathbf{1}_N^\top Q$, $s = t \oslash c$, $Q \leftarrow Q \odot s$,
where $\mathbf{1}_C$ and $\mathbf{1}_N$ are all-ones vectors, $t$ encodes the target
column marginals (e.g., $t = N\pi$ for a prior $\pi$), and $\oslash$ / $\odot$ denote
broadcast division / multiplication.

Each reduction and elementwise pass touches all $NC$ entries, so one iteration costs
$\Theta(NC)$ scalar operations.
Over $K$ iterations, the total complexity is
\begin{equation*}
    \mathcal{O}(KNC).
\end{equation*}
These operations act on the prediction matrix and do not require another forward or backward pass through the model.

\paragraph{Memory footprint.}
The memory cost is dominated by storing the logit queue $Q_{\mathrm{mem}}$, i.e., $NC$
floating-point entries, giving $\mathcal{O}(NC)$ memory.
In our ImageNet experiments we use $C{=}1000$ classes and a window
$N{\approx}1000$ (e.g., $N{=}1024$), so $Q_{\mathrm{mem}}$ contains about $10^6$ entries:
$\approx 2$\,MB in fp16 (or $\approx 4$\,MB in fp32), plus negligible row/column
buffers.
For CIFAR-10/100, where $C \in \{10,100\}$, both computation and memory scale linearly
with $C$ and are correspondingly smaller.

\paragraph{Compute relative to the backbone.}
For the window size and iteration count in Tab.~\ref{tab:scalar_overhead}, the estimated compute cost is below $1\%$ of a ViT-B/16 forward pass. We report measured memory overhead below.

\begin{table}[t]
\centering
\scriptsize
\setlength{\tabcolsep}{4pt}
\caption{\textbf{Backbone vs.\ SCALAR module overhead.}
Compute is per forward pass for ViT-B/16 and per Sinkhorn-based CTTA update
for SCALAR.
SCALAR assumes $C = 1000$ classes (ImageNet), window size $N = 1024$, and
$K = 5$ Sinkhorn iterations, as used in our experiments.
For CIFAR-10/100 ($C\in\{10,100\}$) the overhead is another $10$--$100\times$ smaller.}
\begin{tabular}{lcc}
\toprule
Component & Compute (GFLOPs) & Memory (MB) \\
\midrule
ViT-B/16 backbone (forward) & $\approx 17.5$ & $\approx 330.3$ \\
SCALAR (ImageNet, FP32) & $\approx 0.02$ & $\approx 4$ (fp32 queue) \\
\bottomrule
\end{tabular}
\label{tab:scalar_overhead}
\end{table}

\noindent\textbf{Measured overhead.}
On ImageNet-C ($C{=}1000$, window=1024) with micro-batch size $B{=}8$ on an NVIDIA RTX 3080,
per-step runtime is unchanged within measurement noise (0.1063s $\rightarrow$ 0.1062s)
and peak GPU memory increases by 10MB (\texttt{peak\_alloc}) / 20MB (\texttt{peak\_reserved}).
These measurements use our FP32 implementation, which includes redundant softmax operations. Removing them may reduce the overhead further.
\section{Construction of Skewed Streams}
\label{app:skew}

We construct label-skewed ImageNet-C streams by controlling the class composition of each corruption segment.
The stream stays skewed over extended intervals, allowing prediction biases to accumulate during adaptation.

For each corruption segment, we form a head--tail class split and resample examples according to this split. Head-class examples are oversampled with replacement to a fixed number of samples per head class, while tail classes are subsampled by keeping only a small number of examples per tail class. In our experiments, tail classes are subsampled to one example per class. This preserves the corruption sequence while making head classes more frequent. We generate the sample indices deterministically and reuse them across methods.

\paragraph{Fixed vs.\ dynamic label skew.}
In the \emph{fixed-skew} setting, the skew strength and head-set size are fixed across corruption segments. Specifically, we use 50 head classes and 950 tail classes, oversampling each head class to 100 samples while keeping one example per tail class. The dominant class identities are selected deterministically for each segment, so all methods are evaluated on the same generated stream.

In the \emph{dynamic-skew} setting, the number of head classes and samples per head class vary across corruption segments. We use a cyclic schedule in which the number of head classes follows
\[
(30, 150, 20, 200, 40, 100),
\]
and the corresponding number of samples per head class follows
\[
(170, 33, 250, 25, 125, 50).
\]
Tail classes are again subsampled to one example per class. This changes the per-class head--tail ratio over time, from 25:1 up to 250:1, while keeping the aggregate head mass roughly comparable across segments. Class identities are deterministically reshuffled across segments, and once the schedule is exhausted, it is repeated from the beginning.

Both streams change which classes dominate. The dynamic stream also changes how many classes dominate and how much more often they appear.

\paragraph{Collapse criterion.}
We count a run as collapsed when its post-stream clean error exceeds 95\%.

\paragraph{Oracle prior.}
SCALAR-C uses the empirical class distribution of each corruption segment as its target prior $\pi$, recomputed for every segment from the generated stream. This oracle is used only as a reference; no sample-level labels enter the adaptation procedure.

\paragraph{Long-horizon persistent skew.}
For the five-pass experiment in Tab.~\ref{tab:long_horizon_persistent_skew}, we use the same 50 dominant classes throughout the entire stream. Within each corruption segment, each dominant class contributes 100 samples and each of the remaining 950 classes contributes one sample. A pass consists of all 15 ImageNet-C corruptions processed sequentially, with no model reset between corruptions or passes; each pass contains approximately 89K samples and the full stream approximately 446K samples. Unlike the fixed-skew setting, where the dominant class identities change across segments, here the same dominant set persists throughout. The oracle prior in this experiment is the aggregate class distribution of the stream.

\paragraph{True-subset stream.}
To isolate persistent class absence, we construct a stricter variant in which only the same 50 ImageNet classes appear throughout the stream, and the remaining 950 classes are completely absent. The subset is held fixed across all corruption segments and five consecutive passes, again without model reset. The mismatch between the uniform target and the stream prior therefore persists across all five passes.


\end{document}